\documentclass[conference]{IEEEtran}

\usepackage{amsmath,amssymb,amsfonts}
\usepackage{algorithm}
\usepackage{algorithmic}
\usepackage{booktabs}
\usepackage{multirow}
\usepackage{array}
\usepackage{graphicx}
\usepackage{xcolor}
\usepackage{cite}
\usepackage{url}
\usepackage{colortbl}
\usepackage{tikz}
\usepackage{enumerate}
\usetikzlibrary{arrows.meta, positioning, calc}
\begin{document}

\title{DE-2LS: Differential Evolution with Late-Stage local-search for Unconstrained Single-Objective Numerical Optimization}

\author{\IEEEauthorblockN{Dikshit Chauhan}
\IEEEauthorblockA{Department of Electrical and Computer Engineering, \\National University of Singapore\\
Email: dikshitchauhan608@gmail.com}}

\maketitle

\begin{abstract}
Unconstrained single-objective numerical optimization requires a careful balance among global exploration, late-stage exploitation, and function-evaluation efficiency. This paper presents DE-2LS, a late-stage, local-search-enhanced differential evolution framework built on RDEx for unconstrained single-objective optimization with variable bounds. The proposed method preserves the original RDEx evolutionary search engine and introduces two conservative refinements: a smoothed exploitation-biased branch-rate update in the late search stage and a guarded coordinate-pattern local-search that serves as a budget-aware refinement mechanism. Since the considered setting is unconstrained apart from variable bounds, all selection and local-search acceptance decisions are based solely on objective values. To determine the final algorithm configuration, we conduct a staged ablation study by testing multiple settings of the EB-rate smoothing mechanism, the initial EB-rate, the standard-branch Gaussian sampling scale, the selection-pressure parameters, and the local-search coefficients. The final configuration is selected using a U-score-based evaluation that jointly reflects solution quality and convergence speed. Experimental results show that DE-2LS consistently improves the original RDEx in direct head-to-head comparison. In particular, DE-2LS increases the U-score from $33602.0$ to $37448.0$, corresponding to an improvement of $11.45\%$. Moreover, compared with several competitive and IEEE CEC-winning algorithms, DE-2LS achieves the best overall U-score of $178966.5$, outperforming the others by $34.43\%$. These results show that a carefully designed late-stage local-search strategy can improve both convergence speed and the final objective quality of the algorithm. The source code of DE-2LS is available at \url{https://github.com/ChauhanDikshit?tab=repositories}.
\end{abstract}

\begin{IEEEkeywords}
Differential evolution, Unconstrained optimization, Bound-constrained optimization, CEC competition, Local-search, U-score.
\end{IEEEkeywords}

\section{Introduction}
\IEEEPARstart{D}{ifferential} evolution (DE) is one of the most widely used population-based algorithms for continuous black-box optimization. Since its introduction by Storn and Price \cite{Storn1997DE}, DE has attracted sustained attention due to its simple mutation-crossover-selection cycle, a small number of control parameters, and strong empirical performance on multimodal and nonseparable landscapes \cite{chauhan2025advancements}. In CEC-style unconstrained single-objective optimization problems (UCOPs), the algorithm is typically required to minimize the objective value under a fixed function-evaluation budget while keeping each decision variable within its prescribed bounds.

The IEEE CEC numerical optimization competitions have played an important role in shaping the development of competitive DE variants. Over time, these methods have moved beyond fixed parameter settings and single mutation rules by incorporating success-history adaptation, ensemble mutation strategies, archive-assisted search, population-size reduction, restart mechanisms, covariance information, ranking-based selection pressure, and local-search components \cite{Zhang2009JADE,Tanabe2013SHADE,6900380,Brest2017jSO,Mohamed2017LSHADESPACMA,Sallam2020IMODE,Stanovov2021NLSHADERSP,Stanovov2024LSRTDE,Chauhan2024mLSHADERL,Tao2026RDExSOP}. In the unconstrained setting considered here, there are no additional inequality or equality constraints beyond variable bounds. Therefore, the main algorithmic challenge is to improve objective-error convergence without wasting evaluations on premature or overly aggressive exploitation.

Table~\ref{tab:cec_de_history} summarizes representative DE-related methods and their main mechanisms in CEC-style single-objective optimization. The purpose of the table is not to compare exact scores across different competition tracks, since benchmark suites and scoring protocols have changed over time, but rather to illustrate the methodological trend toward adaptive parameter control, stronger ranking pressure, population reduction, and late-stage exploitation.

\begin{table*}[!t]
\centering
\caption{Representative DE variants and mechanisms for CEC-style single-objective numerical optimization with variable bounds.}
\label{tab:cec_de_history}
\renewcommand{\arraystretch}{1.15}
\begin{tabular}{p{0.08\textwidth}p{0.17\textwidth}p{0.25\textwidth}p{0.34\textwidth}p{0.09\textwidth}}
\toprule
Year & Representative algorithm & UCOP context & Main mechanisms & Ref. \\
\midrule
1997 & DE & Continuous global optimization & Difference-vector mutation, crossover, and greedy objective-based selection & \cite{Storn1997DE} \\
2009 & JADE & Adaptive DE for numerical optimization & Current-to-pbest mutation, optional external archive, and adaptive $F$ and $CR$ control & \cite{Zhang2009JADE} \\
2013 & SHADE & Real-parameter single-objective optimization & Success-history memory for adaptive control parameters & \cite{Tanabe2013SHADE} \\
2014 & L-SHADE & CEC 2014 real-parameter optimization & SHADE with linear population-size reduction & \cite{6900380} \\
2017 & jSO & CEC 2017 bound-constrained single-objective optimization & Improved success-history adaptation, external archive, and population reduction & \cite{Brest2017jSO} \\
2017 & LSHADE-SPACMA & CEC 2017 benchmark problems & Hybrid L-SHADE and CMA-ES style search for nonseparable landscapes & \cite{Mohamed2017LSHADESPACMA} \\
2020 & IMODE & CEC 2020 unconstrained problems & Multi-operator DE with adaptive operator selection and population control & \cite{Sallam2020IMODE} \\
2021 & NL-SHADE-RSP & CEC 2021 bound-constrained optimization & Nonlinear population-size reduction and rank-based selective pressure & \cite{Stanovov2021NLSHADERSP} \\
2024 & L-SRTDE/mLSHADE-RL & CEC 2024 single-objective bound-constrained optimization & Success-rate adaptation, ranking pressure, restart, and late-stage exploitation & \cite{Stanovov2024LSRTDE,Chauhan2024mLSHADERL} \\
2025 & RDEx-SOP & CEC 2025 fixed-budget bound-constrained optimization & Reconstructed DE with exploitation-biased branching and U-score-oriented fixed-budget behavior & \cite{Tao2026RDExSOP} \\
\bottomrule
\end{tabular}
\end{table*}

The evaluation criterion is also important in this context. In earlier CEC-style single-objective studies, performance was commonly emphasized through the final objective error under a fixed function-evaluation budget. In the 2024 competition, the organizers explicitly provided a U-score comparison framework to assess both final solution quality and convergence speed. This change is particularly relevant for DE-based algorithms such as RDEx, where late-stage refinement may not only improve the final objective value but also help the search reach competitive error levels earlier. Therefore, in the present work, the proposed DE-2LS variant is designed and analyzed with explicit attention to U-score-oriented performance.

RDEx provides a strong baseline for CEC-style numerical optimization by combining success-history parameter adaptation, ranking-based parent selection, an exploitation-biased branch, and a decreasing population structure. However, once the search has entered a promising basin, population-level evolutionary operators may still be too stochastic or too coarse to polish the incumbent solution efficiently. This limitation becomes especially important under U-score-based evaluation, where both convergence speed and final objective-error quality affect the final ranking.

Replacing the original RDEx operators directly may damage the search behavior that makes RDEx competitive. Therefore, this UCOP variant follows a conservative ``do no harm'' design principle. The RDEx evolutionary search engine is preserved, while a lightweight local-search is introduced only as a late-stage polishing component. Since the problem is unconstrained except for variable bounds, this local-search uses objective-only acceptance.
The main contributions of this manuscript are as follows:
\begin{enumerate}[(i)]
    \item A late-stage local-search enhanced RDEx variant, named \emph{DE-2LS}, is proposed for CEC-style UCOPs.
    
    \item The proposed method introduces two conservative refinements, a late-stage, smoothed, exploitation-biased branch-rate update and a guarded, coordinate-pattern local-search module, into the original RDEx evolutionary framework.
    
    \item A staged ablation study is conducted to determine the final configuration, including the late EB-related and the local-search-related parameters.
    
    \item Extensive experiments are carried out under a U-score-oriented evaluation protocol, including direct head-to-head comparison with RDEx and broader comparison with several competitive and IEEE CEC-winning algorithms.
    
    \item The results show that DE-2LS improves RDEx in both convergence speed and final objective quality, demonstrating that a carefully designed late-stage local-search mechanism can strengthen RDEx without replacing its main evolutionary search engine.
\end{enumerate}

The remainder of this paper is organized as follows. Section~\ref{sec:de} reviews the basic DE framework for UCOPs. Section~\ref{sec:rdex} summarizes the RDEx mechanisms used as the baseline search engine. Section~\ref{sec:proposed} presents the proposed DE-2LS algorithm. Section~\ref{sec:results} reports the experimental setting, the ablation analysis, and the comparative results. Section~\ref{sec:conclusion} concludes the paper.

\section{Differential Evolution for UCOPs}
\label{sec:de}
Differential evolution (DE) is a population-based evolutionary algorithm for real-parameter optimization \cite{Storn1997DE}. Consider the minimization problem $ \min_{\mathbf{x}\in\Omega} f(\mathbf{x}),$ where $\mathbf{x}=[x_1,x_2,\ldots,x_D]^T$ is a $D$-dimensional decision vector and $\Omega=\{\mathbf{x}\mid L_j\leq x_j\leq U_j,\; j=1,\ldots,D\}$
is the bound-constrained search domain. In the UCOP setting considered in this paper, there are no additional inequality or equality constraints. Therefore, solution quality is determined solely by the objective value, or equivalently by the objective error $f(\mathbf{x})-f^\star$.

At generation $g$, DE maintains a population $\{\mathbf{x}_{i}^{(g)}\}_{i=1}^{N_g}$. For each target vector $\mathbf{x}_{i}^{(g)}$, a mutant vector is generated by combining other population members. A classical mutation strategy is
\begin{equation}
    \mathbf{v}_{i}^{(g)}=\mathbf{x}_{r_1}^{(g)}+F(\mathbf{x}_{r_2}^{(g)}-\mathbf{x}_{r_3}^{(g)}),
\end{equation}
where $r_1$, $r_2$, and $r_3$ are mutually distinct indices and $F$ is the scaling factor. The trial vector $\mathbf{u}_{i}^{(g)}$ is then generated by binomial crossover:
\begin{equation}
    u_{j,i}^{(g)}=
    \begin{cases}
    v_{j,i}^{(g)}, & \text{if } rand_j \leq CR \text{ or } j=j_{rand},\\
    x_{j,i}^{(g)}, & \text{otherwise},
    \end{cases}
\end{equation}
where $CR$ is the crossover rate and $j_{rand}$ guarantees that at least one component is inherited from the mutant vector. For UCOPs, selection is objective-only:
\begin{equation}
    \mathbf{x}_{i}^{(g+1)}=
    \begin{cases}
    \mathbf{u}_{i}^{(g)}, & \text{if } f(\mathbf{u}_{i}^{(g)})\leq f(\mathbf{x}_{i}^{(g)}),\\
    \mathbf{x}_{i}^{(g)}, & \text{otherwise}.
    \end{cases}
\end{equation}
If any generated coordinate violates its bound, the implementation repairs or reinitializes that coordinate inside the interval $[L_j,U_j]$. This objective-only DE framework serves as the basis for the RDEx search engine and the proposed DE-2LS refinements.

\section{Reconstructed Differential Evolution for UCOPs}
\label{sec:rdex}
The baseline search engine adopted in this paper is an unconstrained RDEx variant evaluated through a CEC2017-style objective interface. 
\subsection{Population and Evaluation Setting}
For the main experiments, the decision dimension is set to $D=30$, and the maximum number of function evaluations is $10000D=300000$. The active front size is initialized as $18D$, while the population array is allocated with twice this capacity so that accepted offspring can be stored before compaction. The active front is gradually reduced to four individuals according to
\begin{equation}
    N_g=
    \left\lfloor
    N_0+\frac{4-N_0}{MaxFEs}\,FEs
    \right\rfloor,
\end{equation}
where $N_0$ is the initial front size. This schedule preserves diversity during the early search stage and gradually increases selection pressure as the evaluation budget is consumed.

The experimental protocol uses 25 independent runs with fixed run seeds generated from the master seed \texttt{20260417u}. Objective-error trajectories are recorded at 1000 checkpoints, while the last entry of the result array stores the first function-evaluation count at which the target error is reached. The reported final error is $ e(\mathbf{x})=f(\mathbf{x})-f^\star,$
where the CEC optimum is represented in the implementation as $f^\star=100\times func\_num$.

\subsection{Success-History Parameter Adaptation}
RDEx uses a success-history memory of size five for both $F$ and $CR$, and both memories are initialized to one. Successful parameter values are stored and reused in later generations. The memory update follows a weighted Lehmer-type averaging rule, with weights derived from the corresponding objective improvements. In the standard branch, the center of the scaling-factor distribution at generation $g$ is computed as
\begin{equation}
    \mu_F^{(g)}=0.4+0.25\tanh\!\left(5\,SR^{(g)}\right),
\end{equation}
where $SR^{(g)}$ denotes the current success rate at generation $g$. The scaling factor $F_i^{(g)}$ is then sampled from a truncated Gaussian distribution centered at $\mu_F^{(g)}$, with the sampling scale specified later by the selected algorithm configuration.

\subsection{Ranking-Based Parent Selection}

RDEx maintains ranked index lists based on objective values so that better solutions receive stronger selection probability. In the standard branch, the trial vector at generation $g$ is generated as
\begin{equation}
\begin{split}
    \mathbf{u}_{i}^{(g)}
    = \mathbf{x}_{i}^{(g)}
    + F_i^{(g)}(\mathbf{x}_{p}^{(g)}-\mathbf{x}_{i}^{(g)}) 
    + F_i^{(g)}(\mathbf{x}_{r_1}^{(g,\mathrm{front})}-\mathbf{x}_{r_2}^{(g)}),
\end{split}
\end{equation}
where $\mathbf{x}_{p}^{(g)}$ is sampled from a restricted better subset, $\mathbf{x}_{r_1}^{(g,\mathrm{front})}$ is sampled from the active front, and $\mathbf{x}_{r_2}^{(g)}$ is sampled from the current population. The size of the better subset is controlled by
\begin{equation}
    psize^{(g)}=\max\!\left\{2,\left\lfloor N_g \cdot \xi \cdot \exp\!\left(-k\,SR^{(g)}\right)\right\rfloor\right\},
\end{equation}
where $\xi$ and $k$ are positive coefficients that determine the basic scale of the candidate elite set and its decay with respect to the success rate. Thus, a higher success rate leads to a smaller parent subset and stronger exploitation pressure.

\subsection{Exploitation-Biased Branch}
In addition to the standard branch, RDEx employs an exploitation-biased (EB) branch. In this branch, selected candidate vectors are ordered according to objective value, and the trial vector at generation $g$ is generated from the best, medium, and worst ordered components as
\begin{equation}
    \mathbf{u}_{i}^{(g)}
    =\mathbf{x}_{i}^{(g)}
    +F_i^{(g)}(\mathbf{x}^{g,best}-\mathbf{x}_{i}^{(g)})
    +F_i^{(g)}(\mathbf{x}^{(g,medium)}-\mathbf{x}^{(g,worst)}).
\end{equation}
This branch is intended to strengthen convergence pressure, especially in the later stage of the search. Its selection probability is updated based on the relative improvements produced by the EB and standard branches. The RDEx framework, therefore, combines success-history adaptation, ranking-based parent selection, and branch-level exploitation control, which together form the baseline search engine used in the proposed DE-2LS method. The final parameter values adopted in DE-2LS are introduced later in Section~\ref{sec:proposed} and validated in the ablation analysis.

\section{Proposed DE-2LS Algorithm for UCOPs}
\label{sec:proposed}
This section presents the proposed DE-2LS algorithm for UCOPs. DE-2LS is developed by enhancing the RDEx framework with two late-stage exploitation mechanisms: a smoothed exploitation-biased branch-rate update and a guarded coordinate-pattern local-search module. The main goal is to preserve RDEx's fast global search behavior while improving late-stage refinement and the overall objective-error trajectory.

\subsection{Design Motivation}
RDEx has shown strong performance in real-parameter optimization due to its DE-based search structure, success-history adaptation, exploitation-biased branching mechanism, and decreasing population strategy. However, in the late stage of the search, the population may already lie in a promising basin where additional coordinate-level refinement can further improve the incumbent solution. DE-2LS addresses this issue by introducing a limited-budget local-search component around the current global-best solution.

The design follows two conservative principles. First, the original RDEx evolutionary engine remains the primary search mechanism throughout the run. Second, the local-search module is activated only in the late stage and is strictly budget-controlled. Therefore, DE-2LS does not replace RDEx; instead, it augments RDEx with a lightweight polishing mechanism intended to improve convergence speed and final objective accuracy.

\subsection{Objective-Only Selection Rule}
Since the considered UCOP setting involves only variable bounds, all comparisons in DE-2LS are based solely on objective values. For a candidate solution $\mathbf{x}^{new}$ and an incumbent solution $\mathbf{x}^{old}$, the objective-only acceptance rule is
\begin{equation}
    \mathbf{x}^{new} \prec \mathbf{x}^{old}
    \quad \Longleftrightarrow \quad
    f(\mathbf{x}^{new}) < f(\mathbf{x}^{old}).
\end{equation}
This rule is used consistently to update the global best, accept local-search improvements, and reinject improved solutions into the population.

\subsection{Late-Stage EB-Rate Smoothing}
DE-2LS incorporates a late-stage smoothed exploitation-biased branch-rate mechanism. Let
\begin{equation}
    \eta^{(g)}=\frac{FEs}{MaxFEs}
    \label{eq:progress}
\end{equation}
denote the consumed function-evaluation ratio at generation $g$. Before the late stage, the EB-rate follows the original branch-improvement rule of RDEx. After $\eta^{(g)} \geq \texttt{EB\_SS}$, if both the standard branch and the exploitation-biased branch produce nonzero successful improvement in the current generation, the EB rate is updated as
\begin{equation}
    \rho_{EB}^{(g+1)}
    =
    \texttt{EB\_OW}\,\rho_{EB}^{(g)}
    +
    \texttt{EB\_NW}\,\rho_{EB,\mathrm{raw}}^{(g+1)},
    \label{eq:new EB rate}
\end{equation}
where $\rho_{EB}^{(g)}$ is the previous EB rate and $\rho_{EB,\mathrm{raw}}^{(g+1)}$ is the current branch-improvement-based estimate. This smoothing rule reduces abrupt fluctuations in the branch allocation during the late stage and helps stabilize exploitation around promising regions. If one of the two branches does not contribute any successful improvement in the current generation, the EB rate is reset to its initial value \texttt{EB\_IR}.

\subsection{Guarded Coordinate-Pattern local-search}
\label{subsec:LS}
The second enhancement in DE-2LS is a coordinate-pattern local-search (LS) module centered on the current global-best solution. Let $\mathbf{x}^{b}$ denote the current global best. The LS procedure starts from $\mathbf{x}^{b}$ and explores one coordinate direction at a time. For coordinate $d$, the initial step size is defined as
\begin{equation}
    \Delta_d^{(0)}=\texttt{LS\_ISR}\,(U_d-L_d),
\end{equation}
where $U_d$ and $L_d$ are the upper and lower bounds of the $d$th variable, respectively, and \texttt{LS\_ISR} is the initial step-rate parameter.

For each coordinate $d$, two trial moves are generated:
\begin{equation}
    x_{d}^{+}=x_{d}^{b}+\Delta_d,
    \qquad
    x_{d}^{-}=x_{d}^{b}-\Delta_d.
\end{equation}
If a generated coordinate violates its bound, it is repaired by moving it halfway between the current coordinate value and the violated bound. The LS procedure follows a first-improvement strategy: once a better candidate is found, the LS center is updated immediately, and the coordinate search continues from the improved point. If no coordinate direction yields improvement, the step size is reduced as:
\begin{equation}
    \Delta_d \leftarrow \frac{\Delta_d}{2}.
\end{equation}
The LS process stops when one of the following conditions is met: the LS evaluation budget is exhausted, the overall evaluation budget is exhausted, the maximum number of LS calls is reached, or the current step size becomes smaller than the minimum allowed step.

\subsection{Triggering Logic and Population Injection}
The LS module is activated only in the late stage of the search. In the final DE-2LS setting, LS is allowed after \texttt{LS\_SR} of the evaluation budget. A regular LS call is triggered when the global-best solution has stagnated for at least \texttt{LS\_SG} generations. In addition, a forced very-late LS opportunity is allowed after \texttt{LS\_FSR} of the budget if the LS call limit has not yet been reached. This design ensures that LS is applied only after the evolutionary search has sufficiently explored the search space.

If the LS procedure finds an improved solution, the global best is updated immediately. The improved solution is then injected back into both the active front and the current population by replacing the worst objective-value member whenever the injected solution is better. The success-history memory is not reset, and the population is not restarted. Therefore, the LS module serves as a lightweight refinement step while preserving RDEx's adaptive search state.

\subsection{DE-2LS Parameter Setting}
The final parameter setting of DE-2LS is summarized in Table~\ref{tab:de2ls_params}. The parameters \texttt{EB\_SS}, \texttt{EB\_OW}, \texttt{EB\_NW}, and \texttt{EB\_IR} control the late EB-rate smoothing mechanism; \texttt{EB\_SF} denotes the Gaussian scale used for standard-branch $F$ sampling; \texttt{EB\_XI} and \texttt{EB\_K} define the standard-branch elite-set size rule; and \texttt{EB\_P2} controls the EB-branch elite-subset coefficient. The LS parameters \texttt{LS\_SR}, \texttt{LS\_FSR}, \texttt{LS\_SG}, \texttt{LS\_BR}, \texttt{LS\_MC}, \texttt{LS\_ISR}, and \texttt{LS\_MSR} denote the LS start rate, forced start rate, stagnation threshold, LS budget ratio, maximum number of LS calls, initial step-rate, and minimum step-rate, respectively. Their final values are selected through the ablation analysis in Section~\ref{sec:results}. Algorithm~\ref{alg:de2ls} summarizes the overall DE-2LS framework.
\begin{table}[!h]
\centering
\caption{Final parameter setting of the proposed DE-2LS algorithm.}
\label{tab:de2ls_params}
\renewcommand{\arraystretch}{1.15}
\begin{tabular}{lll}
\hline
Parameter & Value & Meaning \\
\hline
\multicolumn{3}{l}{\textit{Pre-LS RDEx parameters}} \\
\texttt{EB\_SS}  & $0.75$   & Start late EB-rate smoothing after $75\%$ FEs \\
\texttt{EB\_OW}  & $0.65$   & Weight assigned to the previous EB rate \\
\texttt{EB\_NW}  & $0.35$   & Weight assigned to the current EB estimate \\
\texttt{EB\_IR}  & $0.70$   & Initial/default EB-rate value \\
\texttt{EB\_SF}  & $0.025$  & Gaussian scale for standard-branch $F$ sampling \\
\texttt{EB\_XI}  & $0.75$   & Base coefficient of the standard-branch elite-set size \\
\texttt{EB\_K}   & $7.5$    & Exponential decay coefficient for elite-set shrinkage \\
\texttt{EB\_P2}  & $0.17$   & EB-branch elite-subset coefficient \\
\hline
\multicolumn{3}{l}{\textit{Local-search parameters}} \\
\texttt{LS\_SR}  & $0.82$   & Allow LS after $82\%$ FEs \\
\texttt{LS\_FSR} & $0.93$   & Force very-late LS after $93\%$ FEs \\
\texttt{LS\_SG}  & $6$      & Global-best stagnation threshold \\
\texttt{LS\_BR}  & $0.01$   & LS budget ratio per call \\
\texttt{LS\_MC}  & $2$      & Maximum number of LS calls \\
\texttt{LS\_ISR} & $0.015$  & Initial coordinate step as search-width ratio \\
\texttt{LS\_MSR} & $10^{-8}$ & Minimum coordinate step ratio \\
\hline
\end{tabular}
\end{table}

\begin{algorithm}[!h]
\caption{DE-2LS for UCOPs}
\label{alg:de2ls}
\begin{algorithmic}[1]
\STATE Initialize the RDEx population, objective values, memories, active front, and control parameters.
\STATE Set the LS call counter to zero and initialize the global-best solution.
\WHILE{the function-evaluation budget is not exhausted}
    \STATE Generate trial vectors using the standard and exploitation-biased RDEx branches.
    \STATE Repair bound violations and evaluate objective values.
    \STATE Accept trial vectors according to objective-only comparison.
    \STATE Update the global-best solution using objective-only comparison.
    \STATE Update success-history memories and branch-improvement statistics.
    \STATE Compute the progress ratio using Eq.~\eqref{eq:progress}.
    \IF{$\eta^{(g)} \geq \texttt{EB\_SS}$ and both branches have nonzero successful improvement}
        \STATE Smooth the EB rate using Eq.~\eqref{eq:new EB rate}.
    \ELSE
        \STATE Update or reset the EB rate according to the original RDEx branch rule.
    \ENDIF
    \STATE Reduce the active front size according to the RDEx schedule.
    \STATE Update the global-best stagnation counter.
    \IF{$\eta^{(g)} \geq \texttt{LS\_SR}$ and stagnation $\geq \texttt{LS\_SG}$ and the LS call limit is not reached}
        \STATE Run coordinate-pattern local-search around the current global best.
    \ELSIF{$\eta^{(g)} \geq \texttt{LS\_FSR}$ and the LS call limit is not reached}
        \STATE Run one forced very-late local-search call.
    \ENDIF
    \IF{local-search returns a lower objective value}
        \STATE Update the global best.
        \STATE Inject the improved solution into the population and active front.
    \ENDIF
\ENDWHILE
\STATE Return the final best solution and the recorded objective-error curve.
\end{algorithmic}
\end{algorithm}

\section{Results and Discussion}
\label{sec:results}

\subsection{Experimental Setting}
\label{subsec:experimental_setting}
The experiments are conducted on UCOPs under objective-only function-evaluation accounting. Following the CEC-style single-objective evaluation protocol\footnote{\url{https://github.com/P-N-Suganthan/2026-CEC}}, the decision dimension is set to $D=30$, and the maximum number of function evaluations is $MaxFEs = 10000\times D$. For each test function and each algorithm, 25 independent runs are performed using fixed random seeds generated from a master seed to ensure reproducibility. The objective-error value is recorded at fixed checkpoints during the search. The final objective error is computed as $f(\mathbf{x})-f^\star$, where $f^\star$ is the known optimum of the corresponding test function. If the computed error is less than or equal to $10^{-8}$, it is clipped to zero.

The compared algorithms are DE-2LS, the original RDEx baseline without the proposed local-search module \cite{Tao2026RDExSOP}, and the competitive algorithms LSRTDE \cite{Stanovov2024LSRTDE}, BlockEA, mLSHADE-RL \cite{Chauhan2024mLSHADERL}, jSO \cite{Brest2017jSO}, and IEACOP \cite{10611920}. All algorithms use the same stopping criterion and the same evaluation budget.

Performance is evaluated using the pairwise U-score procedure for UCOPs. For each pair of runs, the accuracy component compares their final objective errors, while the speed component compares how quickly each run reaches the pairwise reference error level. Therefore, the U-score jointly reflects final solution quality and convergence speed.

For each algorithm, the overall U-score is obtained by summing the function-wise U-scores over all benchmark functions. A larger U-score indicates better overall performance. When algorithm $a$ is compared against a baseline algorithm $b$, the relative gain is computed as
\begin{equation}
\text{Relative Gain}(a \mid b)
=
\frac{\mathcal{U}_a-\mathcal{U}_b}{\mathcal{U}_b}\times 100,
\label{eq:relative_gain}
\end{equation}
where $\mathcal{U}_a$ and $\mathcal{U}_b$ are the corresponding overall U-scores. A positive value indicates that algorithm $a$ outperforms the baseline.

\subsubsection{Procedure for U-score}
\label{subsubsec:procedure_uscore_ucop}
\begin{figure*}[!t]
\centering
\begin{tikzpicture}[
    font=\small,
    >=Latex,
    box/.style={draw, rounded corners, align=center, minimum height=0.5cm, minimum width=2cm},
    arr/.style={->, thick}
]

\node[anchor=west] at (0,4.2) {\textbf{(a) Objective-error trajectories}};

\draw[->] (0,0) -- (5.2,0) node[below] {Checkpoints/FEs};
\draw[->] (0,0) -- (0,3.6) node[above] {Objective error};

\draw[thick] plot[smooth] coordinates {(0.2,3.2) (1.0,2.3) (2.0,1.45) (3.2,0.85) (4.7,0.45)};
\draw[thick,dashed] plot[smooth] coordinates {(0.2,3.35) (1.0,2.75) (2.0,2.05) (3.2,1.45) (4.7,1.05)};

\fill (4.7,0.45) circle (2pt);
\fill (4.7,1.05) circle (2pt);

\node[right] at (4.75,0.45) {$E_x^\star$};
\node[right] at (4.75,1.05) {$E_y^\star$};

\node at (1.3,1.5) {Run $x$};
\node at (1.6,2.75) {Run $y$};

\node[anchor=west] at (6.2,4.2) {\textbf{(b) Pairwise scoring}};

\draw[->] (6.2,0) -- (11.4,0);
\draw[->] (6.2,0) -- (6.2,3.6) node[above] {Objective error};

\draw[thick] plot[smooth] coordinates {(6.4,3.2) (7.2,2.3) (8.2,1.45) (9.4,0.85) (10.9,0.45)};
\draw[thick,dashed] plot[smooth] coordinates {(6.4,3.35) (7.2,2.75) (8.2,2.05) (9.4,1.45) (10.9,1.05)};

\draw[dotted, thick] (6.2,1.05) -- (11,1.05);
\node[above] at (11,1.05) {$\theta_E$};
\node[above] at (10,2) {$\theta_E=\max(E_x^\star,E_y^\star)$};

\draw[dotted, thick] (8.9,0) -- (8.9,1.05);
\draw[dotted, thick] (10.9,0) -- (10.9,1.05);

\node[below] at (8.9,0) {$\tau_x$};
\node[below] at (10.9,0) {$\tau_y$};

\fill (10.9,0.45) circle (2pt);
\fill (10.9,1.05) circle (2pt);

\node[box, minimum width=4.4cm, fill=purple!20] at (2,-1.5)
{Accuracy: compare $E_x^\star$ and $E_y^\star$\\
smaller final error gets point};

\node[box, minimum width=4.4cm, fill=cyan!20] at (8.8,-1.5)
{Speed: compare $\tau_x$ and $\tau_y$\\
earlier reaching time gets point};

\node[anchor=west] at (12.4,4.2) {\textbf{(c) Aggregation}};

\node[box, fill=red!20] (allruns) at (14.2,3.1) {All runs from\\all algorithms};
\node[box, fill=blue!20] (pairs) at (14.2,1.9) {Pairwise trial\\comparisons};
\node[box, fill=green!20] (ascore) at (12.7,0.6) {Accuracy score \bf (a)};
\node[box, fill=green!20] (sscore) at (15.7,0.6) {Speed score \bf (b)};
\node[box, minimum width=4.2cm, fill=blue!20] (uscore) at (14.2,-0.9)
{$\mathcal{U}_a^{(f)}=\mathcal{A}_a^{(f)}+\mathcal{S}_a^{(f)}$};
\node[box, minimum width=4.2cm, fill=red!20] (total) at (14.2,-2.25)
{$\mathcal{U}_a=\sum_f \mathcal{U}_a^{(f)}$};

\draw[arr] (allruns) -- (pairs);
\draw[arr] (pairs) -- (ascore);
\draw[arr] (pairs) -- (sscore);
\draw[arr] (ascore) -- (uscore);
\draw[arr] (sscore) -- (uscore);
\draw[arr] (uscore) -- (total);

\end{tikzpicture}
\caption{Illustration of the pairwise U-score procedure for UCOPs. 
(a) Each run records the best-so-far objective-error trajectory at fixed checkpoints. 
(b) For a pair of runs, the accuracy component compares the final objective errors, while the speed component compares the first time each run reaches the pairwise reference error level $\theta_E=\max(E_x^\star,E_y^\star)$. 
(c) Pairwise accuracy and speed scores are accumulated over all trial pairs to obtain the function-wise U-score and then summed over all functions.}
\label{fig:uscore_ucop_procedure}
\end{figure*}
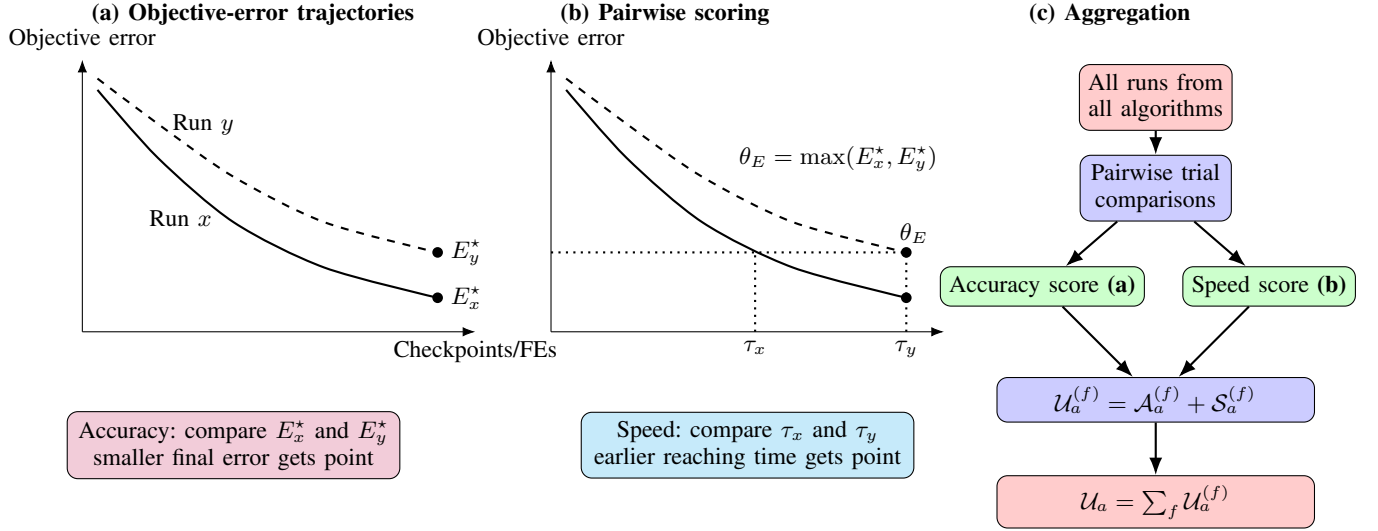

For the U-score calculation, we follow the pairwise scoring procedure for UCOPs. Since UCOPs do not involve constraint violations, the scoring is based only on the objective-error trajectory. Each run stores the best-so-far objective error, denoted by \texttt{Min\_EV}, after initialization and then at fixed evaluation intervals. As specified in the technical report, the objective error is recorded every $10D$ evaluations until \texttt{MaxFEs}. Since $D=30$ and \texttt{MaxFEs}$=10000D$, each run contains 1000 saved points, including the initial point. The procedure of U-score calculation is visually illustrated in Fig.~\ref{fig:uscore_ucop_procedure}.

Let $E_{a,r}^{(f)}(t)$ denote the saved \texttt{Min\_EV} value of algorithm $a$, run $r$, function $f$, and sampling point $t$. The objective error is defined as
\begin{equation}
E_{a,r}^{(f)}(t)=f\!\left(\mathbf{x}_{a,r}^{best}(t)\right)-f^\star,
\end{equation}
where $\mathbf{x}_{a,r}^{best}(t)$ is the best-so-far solution at sampling point $t$, and $f^\star$ is the known optimum of the corresponding function. Following the numerical precision setting, objective-error values smaller than or equal to $\epsilon_{\mathrm{EV}}=10^{-8}$ are set to zero.

Let $m$ be the number of compared algorithms and $n$ the number of independent runs. For each function, all $mn$ trials are compared pairwise. The number of unordered trial pairs is
\begin{equation}
    N_{\mathrm{pair}}=\frac{mn(mn-1)}{2}.
\end{equation}
For two trials $x$ and $y$, let $E_x^\star$ and $E_y^\star$ denote their final saved objective errors at \texttt{MaxFEs}. Since a smaller objective error is better, the comparison function is defined as
\begin{equation}
\label{eq:rho_ucop}
\rho(u,v;\epsilon)=
\begin{cases}
1, & u < v-\epsilon,\\
0.5, & |u-v|\leq \epsilon,\\
0, & u > v+\epsilon.
\end{cases}
\end{equation}
Here, $\rho(u,v;\epsilon)$ denotes the point awarded to the first trial when comparing $u$ and $v$.

The pairwise accuracy point awarded to trial $x$ against trial $y$ is
\begin{equation}
\label{eq:pairwise_accuracy_ucop}
A(x,y)=\rho(E_x^\star,E_y^\star;\epsilon_{\mathrm{EV}}).
\end{equation}
Thus, the trial with the smaller final objective error receives one accuracy point. If both final errors are equal within the tolerance $\epsilon_{\mathrm{EV}}$, both trials receive 0.5 points.

The accuracy score of algorithm $a$ on function $f$ is obtained by summing the pairwise accuracy points of its $n$ trials against all other trials:
\begin{equation}
\label{eq:algorithm_accuracy_ucop}
    \mathcal{A}_{a}^{(f)}
    =
    \sum_{x\in \mathcal{T}_{a}^{(f)}}
    \sum_{\substack{y\in \mathcal{T}^{(f)}\\y\neq x}}
    A(x,y),
\end{equation}
where $\mathcal{T}_{a}^{(f)}$ is the set of trials of algorithm $a$ on function $f$, and $\mathcal{T}^{(f)}$ is the set of all trials from all compared algorithms on function $f$.

The speed score is computed by comparing the time needed to reach a pairwise reference error level. For a pair of trials $x$ and $y$, the reference level is defined as
\begin{equation}
\label{eq:theta_error_ucop}
    \theta_E(x,y)=\max(E_x^\star,E_y^\star).
\end{equation}
The first reaching index of trial $x$ with respect to this reference level is then
\begin{equation}
\label{eq:tau_error_ucop}
    \tau_E(x,y)=
    \min \left\{t \mid E_x(t)\leq \theta_E(x,y)+\epsilon_{\mathrm{EV}}\right\}.
\end{equation}
If the reference level is never reached, $\tau_E(x,y)$ is set to infinity.

The pairwise speed point awarded to trial $x$ against trial $y$ is
\begin{equation}
\label{eq:pairwise_speed_ucop}
S(x,y)=\rho(\tau_E(x,y),\tau_E(y,x);0).
\end{equation}
Therefore, the trial that reaches the pairwise reference error level earlier receives one speed point. If both trials reach the reference level at the same sampling point, each receives 0.5 speed points.

The speed score of algorithm $a$ on function $f$ is computed as
\begin{equation}
\label{eq:algorithm_speed_ucop}
    \mathcal{S}_{a}^{(f)}
    =
    \sum_{x\in \mathcal{T}_{a}^{(f)}}
    \sum_{\substack{y\in \mathcal{T}^{(f)}\\y\neq x}}
    S(x,y).
\end{equation}

The U-score of algorithm $a$ on function $f$ is the sum of its accuracy and speed scores:
\begin{equation}
\label{eq:function_uscore_ucop}
    \mathcal{U}_{a}^{(f)}
    =
    \mathcal{A}_{a}^{(f)}+\mathcal{S}_{a}^{(f)}.
\end{equation}
The final U-score over all $K=29$ UCOP functions is
\begin{equation}
\label{eq:final_uscore_ucop}
    \mathcal{U}_{a}
    =
    \sum_{f=1}^{K}
    \mathcal{U}_{a}^{(f)}.
\end{equation}
A larger U-score indicates better overall performance because it reflects both final objective-error quality and convergence speed. In the reported results, the accuracy, speed, and U-score are obtained by summing $\mathcal{A}_{a}^{(f)}$, $\mathcal{S}_{a}^{(f)}$, and $\mathcal{U}_{a}^{(f)}$ over all test functions, respectively.

\subsection{Ablation Analysis}
\label{subsec:ablation}
To determine the final configuration of DE-2LS, we conducted the ablation study in two stages. In the first stage, we optimized the underlying RDEx search dynamics before activating local-search (LS). In the second stage, we fixed the best pre-LS configuration and then analyzed the LS-related parameters. Throughout this study, the variants were compared using accuracy, speed, and U-score.

\subsubsection{Ablation of late EB-rate smoothing}
\label{subsec:ablation_eb_smoothing}
We first examined whether the EB-rate update should be smoothed in the later stage of the search. Let $\rho_{EB,\mathrm{raw}}^{(g+1)}$ denote the raw EB-rate estimate at generation $g+1$, computed from the relative objective improvements of the successful EB and standard-branch trials:
\begin{equation}
\rho_{EB,\mathrm{raw}}^{(g+1)}
=
\frac{\sum_{i \in S_{EB}^{(g)}} \Delta f_i}
{\sum_{i \in S_{EB}^{(g)}} \Delta f_i + \sum_{i \in S_{\mathrm{std}}^{(g)}} \Delta f_i},
\label{eq:eb_raw}
\end{equation}
where $S_{EB}^{(g)}$ and $S_{\mathrm{std}}^{(g)}$ are the sets of successful EB and standard-branch trials, respectively, and $\Delta f_i$ is the corresponding improvement in objective value.

To stabilize this estimate, the EB-rate is updated as
\begin{equation}
\rho_{EB}^{(g+1)} =
\begin{cases}
\rho_{EB,\mathrm{raw}}^{(g+1)}, & \eta < \texttt{EB\_SS}, \\[4pt]
\texttt{EB\_OW}\,\rho_{EB}^{(g)}+\texttt{EB\_NW}\,\rho_{EB,\mathrm{raw}}^{(g+1)}, & \eta \ge \texttt{EB\_SS},
\end{cases}
\label{eq:eb_smooth}
\end{equation}
where \texttt{EB\_SS} is the smoothing start point, \texttt{EB\_OW} is the weight assigned to the previous EB-rate value, and \texttt{EB\_NW} is the weight assigned to the current raw estimate. If one of the two branches does not produce a successful improvement, the EB-rate is reset to \texttt{EB\_IR}.

\begin{table}[!h]
\centering
\caption{Ablation results for late EB-rate smoothing variants.}
\label{tab:ablation_eb_smoothing}
\begin{tabular}{lccc}
\hline
Variant	&	Accuracy	&	Speed	&	U-score	\\\hline
RDEx	&	45211.5	&	45201	&	90412.5	\\
\texttt{EB\_SS73}	&	45142.5	&	44811	&	89953.5	\\
\cellcolor{gray!25}\bf\texttt{EB\_SS75}	&\cellcolor{gray!25}\bf	45758.5	&\cellcolor{gray!25}\bf	45882.5	&\cellcolor{gray!25}\bf	91641	\\
\texttt{EB\_OW80}	&	45082.5	&	44873	&	89955.5	\\
\texttt{EB\_OW50}	&	43555	&	43982.5	&	87537.5	\\\hline
\end{tabular}
\end{table}
Table~\ref{tab:ablation_eb_smoothing} shows that the effect of late EB-rate smoothing depends strongly on both the activation point and the smoothing weights. Among the tested variants, \texttt{EB\_SS75} achieves the best overall performance, with an accuracy score of $45758.5$, a speed score of $45882.5$, and the highest U-score of $91641$. Compared with the \texttt{RDEx} baseline, this corresponds to an absolute U-score improvement of $1228.5$ ($1.36\%$), indicating that activating the smoothing mechanism after $75\%$ of the function evaluations provides a better late-stage balance between refinement and stability.

In contrast, starting the smoothing earlier at \texttt{EB\_SS73} slightly reduces both accuracy and speed, leading to a lower U-score than the baseline. A similar trend is observed for \texttt{EB\_OW80}, which remains relatively stable but still underperforms \texttt{RDEx}. The weakest result is obtained by \texttt{EB\_OW50}, where reducing the carry-over effect too aggressively degrades both accuracy and speed and lowers the U-score to $87537.5$. These results indicate that late EB-rate smoothing is beneficial only when introduced at an appropriate stage and with a moderate level of carryover from the previous EB-rate value. Therefore, the late-smoothing configuration is fixed at \texttt{EB\_SS}$=0.75$, \texttt{EB\_OW}$=0.65$, and \texttt{EB\_NW}$=0.35$.

\subsubsection{Ablation of the initial EB-rate}
\label{subsec:ablation_eb_ir}
After fixing the late-smoothing mechanism, we next studied the initial EB-rate parameter. The EB-rate is initialized as
\begin{equation}
\rho_{EB}^{(0)} = \texttt{EB\_IR},
\label{eq:eb_init}
\end{equation}
where \texttt{EB\_IR} is the default EB-rate value used at the start of the run.
\begin{table}[!h]
\centering
\caption{Ablation results for initial EB-rate variants.}
\label{tab:ablation_eb_ir}
\begin{tabular}{lccc}
\hline
Variant & Accuracy & Speed & U-score \\
\hline
\texttt{EB\_IR068} & 26938.5 & 26931 & 53869.5 \\
\cellcolor{gray!25}\bf\texttt{EB\_IR070} &\cellcolor{gray!25}\bf 27023 &\cellcolor{gray!25}\bf 27012.5 &\cellcolor{gray!25}\bf 54035.5 \\
\texttt{EB\_IR072} & 26513.5 & 26531.5 & 53045 \\
\hline
\end{tabular}
\end{table}
Table~\ref{tab:ablation_eb_ir} indicates that \texttt{EB\_IR070} provides the best overall performance, achieving the highest U-score. Although \texttt{EB\_IR068} remains competitive, its accuracy and speed are both slightly below those of \texttt{EB\_IR070}. Increasing the initial EB rate to \texttt{EB\_IR072} results in clearer deterioration. Therefore, \texttt{EB\_IR}$=0.70$ is retained.

\subsubsection{Ablation of the standard-branch Gaussian scale}
\label{subsec:ablation_eb_sf}

Next, we examined the Gaussian scale used to sample the scaling factor in the standard branch. In the implementation, the scaling factor is sampled from a truncated Gaussian distribution:
\begin{equation}
F^{(g)} \sim \mathcal{N}\!\left(\mu_F^{(g)}, \texttt{EB\_SF}^2\right),
\qquad 0 < F^{(g)} < 1,
\label{eq:eb_sf}
\end{equation}
where $\mu_F^{(g)}$ is the current center of the Gaussian sampling and \texttt{EB\_SF} is its standard deviation.

\begin{table}[!h]
\centering
\caption{Ablation results for standard-branch Gaussian-scale variants.}
\label{tab:ablation_eb_sf}
\begin{tabular}{lccc}
\hline
Variant	&	Accuracy	&	Speed	&	U-score	\\\hline
\texttt{EB\_SF015}	&	35957.5	&	33198	&	69155.5	\\
\texttt{EB\_SF0225}	&	35775.5	&	35530.5	&	71306	\\
\texttt{EB\_SF025}	&	35715.5	&	36618.5	&	72334	\\
\cellcolor{gray!25}\bf\texttt{EB\_SF0275}	&\cellcolor{gray!25}\bf	36101.5	&\cellcolor{gray!25}\bf	38203	&\cellcolor{gray!25}\bf	74304.5 \\
\hline
\end{tabular}
\end{table}
Table~\ref{tab:ablation_eb_sf} shows that the standard-branch Gaussian scale has a clear effect on both convergence speed and overall U-score. Among the tested variants, \texttt{EB\_SF0275} achieves the best overall performance, with an accuracy score of $36101.5$, a speed score of $38203$, and the highest U-score of $74304.5$. Compared with \texttt{EB\_SF025}, this corresponds to an additional U-score gain of $1970.5$ ($2.72\%$), indicating that a slightly larger Gaussian scale provides a stronger balance between exploration and late-stage refinement.

A smaller value, \texttt{EB\_SF015}, is clearly too restrictive, yielding the lowest speed score and the lowest U-score among all tested settings. Increasing the scale to \texttt{EB\_SF0225} improves the speed component and raises the U-score, while \texttt{EB\_SF025} provides a further gain. However, the best result is obtained at \texttt{EB\_SF0275}, which outperforms all smaller settings in both accuracy and speed. These results suggest that, in the present framework, a moderately larger Gaussian sampling scale improves the effectiveness of the standard branch. 

\subsubsection{Ablation of the selection-pressure parameters}
\label{subsec:ablation_xi_k}

After fixing the late-smoothing and Gaussian-scale settings, we further analyzed the selection-pressure parameters $\xi$ and $k$. These parameters jointly determine the size of the standard-branch elite set:
\begin{equation}
p^{(g)} = \max\!\left(2,\left\lfloor N^{(g)} \cdot \xi \cdot \exp\!\left(-k \cdot SR^{(g)}\right)\right\rfloor\right),
\label{eq:ablation_xi_k}
\end{equation}
where $N^{(g)}$ is the current front size and $SR^{(g)}$ denotes the success rate at generation $g$. Here, $\xi$ controls the base scale of the elite set, while $k$ determines how aggressively the elite-set size shrinks with increasing success rate.

\begin{table}[!h]
\centering
\caption{Ablation results for the selection-pressure parameters $\xi$ and $k$.}
\label{tab:ablation_xi_k}
\begin{tabular}{lccc}
\hline
Variant	&	Accuracy	&	Speed	&	U-score	\\\hline
\texttt{EB\_XI075\_K75}	&	52829	&	56108	&	108937	\\
\texttt{EB\_XI072\_K75}	&	52224.5	&	57132.5	&	109357	\\
\cellcolor{gray!25}\bf\texttt{EB\_XI068\_K75}	&	52506.5	&\cellcolor{gray!25}\bf	59585.5	&\cellcolor{gray!25}\bf	112092	\\
\texttt{EB\_XI065\_K65}	&	54131	&	52935	&	107066	\\
\texttt{EB\_XI075\_K65}	&\cellcolor{gray!25}\bf	56192	&	49157	&	105349	\\\hline
\end{tabular}
\end{table}

Table~\ref{tab:ablation_xi_k} shows that the selection-pressure parameters $\xi$ and $k$ are strongly coupled and should be tuned jointly. Among the tested variants, \texttt{EB\_XI068\_K75} achieves the best overall performance, with an accuracy score of $52506.5$, a speed score of $59585.5$, and the highest U-score of $112092$. Compared with \texttt{EB\_XI075\_K75}, this corresponds to an additional U-score gain of $3155.0$ ($2.90\%$), driven mainly by a substantial improvement in the speed component.

A comparison among the three variants with fixed $k=7.5$ shows that reducing $\xi$ improves the overall balance of the method. While \texttt{EB\_XI075\_K75} and \texttt{EB\_XI072\_K75} remain competitive, the best result is obtained by \texttt{EB\_XI068\_K75}, indicating that a smaller elite-set scale is more effective in the present framework. In contrast, deviating from $7.5$ in $k$ degrades performance. Both \texttt{EB\_XI075\_K70} and \texttt{EB\_XI075\_K65} obtain the same lower U-score of $105349$, while \texttt{EB\_XI065\_K65} also remains below the best setting. These results indicate that the decay coefficient $k=7.5$ should be preserved, whereas the elite-set scaling coefficient $\xi$ benefits from a more conservative value.


\subsubsection{Ablation of local-search coefficients}
\label{subsec:ablation_ls}
After fixing the pre-LS RDEx configuration, we integrated the late local-search component and analyzed its parameters in three groups: timing, budget, and step-size control. The timing variants T1-T3 modify when LS is allowed and forced. Specifically, DE-2LS-(T1) represents an earlier and more aggressive LS configuration with \texttt{LS\_SR}$=0.82$, \texttt{LS\_FSR}$=0.93$, and \texttt{LS\_SG}$=6$; DE-2LS-(T2/B2/S2) is the baseline LS setting with \texttt{LS\_SR}$=0.85$, \texttt{LS\_FSR}$=0.95$, and \texttt{LS\_SG}$=8$; and DE-2LS-(T3) delays LS activation with \texttt{LS\_SR}$=0.88$, \texttt{LS\_FSR}$=0.97$, and \texttt{LS\_SG}$=10$.

The budget variants B1-B4 control the computational effort assigned to LS. DE-2LS-(B1) uses a light LS setting with \texttt{LS\_BR}$=0.005$ and \texttt{LS\_MC}$=1$; DE-2LS-(B2) is the baseline setting with \texttt{LS\_BR}$=0.010$ and \texttt{LS\_MC}$=2$; DE-2LS-(B3) moderately increases the LS budget with \texttt{LS\_BR}$=0.015$ and \texttt{LS\_MC}$=2$; and DE-2LS-(B4) uses a heavier exploratory setting with \texttt{LS\_BR}$=0.020$ and \texttt{LS\_MC}$=3$.

Finally, the step-size variants S1-S3 examine the effect of the initial LS step size. DE-2LS-(S1) uses \texttt{LS\_ISR}$=0.010$, DE-2LS-(T2/B2/S2) uses the baseline \texttt{LS\_ISR}$=0.015$, and DE-2LS-(S3) uses \texttt{LS\_ISR}$=0.020$; all three retain \texttt{LS\_MSR}$=10^{-8}$.

\begin{table}[!h]
\centering
\caption{U-score comparison of DE-2LS variants with different local-search parameter settings.}
\label{tab:de2ls_ls_parameter_analysis}
\resizebox{0.5\textwidth}{!}{%
\begin{tabular}{lcccccc}
\hline
Algos & 
Accuracy & 
Speed & 
U-score & 
Rank Sum & 
Avg. Rank & 
Wins \\
\hline
\cellcolor{gray!25}\textbf{DE-2LS-(T1)}       & \cellcolor{gray!25}\textbf{74430.5} & \cellcolor{gray!25}\textbf{77536.5} & \cellcolor{gray!25}\textbf{151967} & \cellcolor{gray!25}\textbf{92}  & \cellcolor{gray!25}\textbf{3.7414} &\cellcolor{gray!25} \textbf{12} \\
DE-2LS-(T3)       & 73782.5 & 72131.5 & 145914 & 129 & 4.2759 & 1 \\
DE-2LS-(S1)       & 71936 & 72689.5 & 144625.5 & 137 & 4.0690 & 4 \\
DE-2LS-(B4)       & 71940.5 & 72099 & 144039.5 & 164 & 4.3103 & 3 \\
DE-2LS-(B1)       & 71524 & 72131.5 & 143656 & 126 & 5.0345 & 3 \\
DE-2LS-(S3)       & 71950 & 70934 & 142884 & 119 & 4.2241 & 3 \\
DE-2LS-(T2/B2/S2) & 71633 & 70787.5 & 142420.5 & 131 & 4.9655 & 2 \\
DE-2LS-(B3)       & 69903 & 68790.5 & 138693.5 & 146 & 5.3793 & 1 \\
\hline
\end{tabular}%
}\begin{flushleft}
\footnotesize \textit{Note:} In this table, a larger U-score, accuracy, speed, and number of wins indicate better performance, whereas a smaller U-score rank sum and average rank indicate better performance. Avg. Rank and Wins are calculated based on the mean fitness values. 
\end{flushleft}
\end{table}
As shown in Table~\ref{tab:de2ls_ls_parameter_analysis}, DE-2LS-(T1) achieves the best overall performance among all tested variants. It obtains the highest U-score of $151967$, the highest speed score of $77536.5$, the lowest U-score rank sum, the best average rank by final mean error, and the largest number of wins. This indicates that activating LS earlier and using a more aggressive stagnation trigger is beneficial for the proposed framework. Compared with the baseline DE-2LS-(T2/B2/S2), DE-2LS-(T1) improves the U-score from $142420.5$ to $151967$.

The delayed timing variant DE-2LS-(T3) also improves over the baseline, but remains clearly below DE-2LS-(T1), indicating that postponing LS weakens the benefit of early refinement. Among the budget variants, DE-2LS-(B4) performs better than the baseline and the lighter B1 setting, suggesting that increasing the LS budget can be useful to some extent. However, DE-2LS-(B3) gives the lowest U-score among all variants, which indicates that excessive or poorly balanced LS effort may consume evaluations without producing proportional gains. For the step-size variants, DE-2LS-(S1), which uses a smaller initial LS step, outperforms both the baseline and the larger-step variant DE-2LS-(S3), suggesting that a conservative LS step is more reliable for local refinement. Overall, the results show that the LS timing parameters have the strongest influence on DE-2LS. Therefore, DE-2LS-(T1) is selected as the final LS configuration for the subsequent experiments, and the detailed parameter settings are shown in Table \ref{tab:de2ls_params}.

\subsection{Head-to-Head Comparison with RDEx}
\label{subsec:comparison_rdex_ucop}
Since DE-2LS is derived from RDEx, the most important assessment is a direct head-to-head comparison between these two algorithms on UCOPs. Table~\ref{tab:de2ls_vs_rdex_ucop_summary} summarizes their accuracy, speed, U-score, U-score rank sum, and average rank by mean final error. In this comparison, the variant previously denoted as \texttt{DE-2LS-(T1)} is reported simply as DE-2LS.

\begin{table}[!h]
\centering
\caption{Head-to-head comparison between DE-2LS and RDEx on UCOPs.}
\label{tab:de2ls_vs_rdex_ucop_summary}
\renewcommand{\arraystretch}{1.12}
\resizebox{\linewidth}{!}{%
\begin{tabular}{lrrrrr}
\hline
Algos & Accuracy &Speed & U-score & 
Rank Sum & Avg. Rank \\
\hline
\cellcolor{gray!25}DE-2LS &\cellcolor{gray!25} \textbf{18275.5} &\cellcolor{gray!25} \textbf{19172.5} &\cellcolor{gray!25} \textbf{37448} &\cellcolor{gray!25} \textbf{38} &\cellcolor{gray!25} \textbf{1.4828}  \\
RDEx   & 17249.5 & 16352.5 & 33602 & 49 & 1.5172  \\
\hline
Relative gain 
& $+5.95\%$ & $+17.25\%$ & $+11.45\%$ & $22.45\%\downarrow$ & $2.27\%\downarrow$ \\
\hline
\end{tabular}%
}\begin{flushleft}
\footnotesize \textit{Note:} Here, $\downarrow$ indicates that a smaller value is better. 
\end{flushleft}
\end{table}

As shown in Table~\ref{tab:de2ls_vs_rdex_ucop_summary}, DE-2LS achieves a higher overall U-score than RDEx, improving the U-score from $33602$ to $37448$, which corresponds to a relative gain of $11.45\%$ according to Eq.~\eqref{eq:relative_gain}. This improvement is obtained from both components of the U-score. In terms of accuracy, DE-2LS improves the score from $17249.5$ to $18275.5$, corresponding to a relative gain of $5.95\%$. The improvement is even more evident in the speed component, where DE-2LS increases the score from $16352.5$ to $19172.5$, corresponding to a relative gain of $17.25\%$.

The rank-based indicators further support this observation. DE-2LS reduces the U-score rank sum from $49$ to $38$, which corresponds to a $22.45\%$ reduction, while the average rank by mean final error improves from $1.5172$ to $1.4828$. Although the improvement in average rank is modest, the gain in U-score is substantial, indicating that the main benefit of DE-2LS lies in improving the balance between convergence speed and final accuracy.

The function-wise results in Table~\ref{tab:function-wise} provide a more detailed view of this behavior. DE-2LS obtains a better U-score on $20$ out of the $29$ UCOP functions, whereas RDEx performs better on the remaining $9$ functions. A closer inspection of the two U-score components shows that DE-2LS is particularly effective in improving convergence speed: it achieves a better speed score on $18$ functions, while RDEx is better on $11$ functions. The improvement is especially clear on functions such as F1, F3, F8, F9, F22, and F28, where DE-2LS obtains substantially larger speed scores than RDEx. In terms of accuracy, DE-2LS also shows an advantage, but the improvement is more moderate. DE-2LS achieves higher accuracy on $16$ functions, RDEx on $9$, and both algorithms are tied on $4$. Overall, these results show that the proposed LS enhancement improves both speed and final solution quality without sacrificing the strong global-search behavior of RDEx.

\begin{table}[!h]
\caption{Function-wise pairwise U-score comparison between DE-2LS and RDEx.}
\label{tab:function-wise}
\resizebox{1\linewidth}{!}{%
\begin{tabular}{c|cc|cc|cc}
\hline
\multirow{2}{*}{Func} & \multicolumn{2}{c|}{Accuracy} & \multicolumn{2}{c|}{Speed} & \multicolumn{2}{c}{U-score} \\
\cline{2-7}
& DE-2LS & RDEx & DE-2LS & RDEx & DE-2LS & RDEx \\
\hline
F1  & 612.5 & 612.5 & \cellcolor{gray!75}909.5 & 315.5 & \cellcolor{gray!75}1522 & 928 \\
F3  & 612.5 & 612.5 & \cellcolor{gray!75}874.5 & 350.5 & \cellcolor{gray!75}1487 & 963 \\
F4  & 437 & \cellcolor{gray!75}788 & 438 & \cellcolor{gray!75}787 & 875 & \cellcolor{gray!75}1575 \\
F5  & \cellcolor{gray!75}661.5 & 563.5 & \cellcolor{gray!75}702.5 & 522.5 & \cellcolor{gray!75}1364 & 1086 \\
F6  & \cellcolor{gray!75}787.5 & 437.5 & 590.5 & \cellcolor{gray!75}634.5 & \cellcolor{gray!75}1378 & 1072 \\
F7  & \cellcolor{gray!75}623 & 602 & \cellcolor{gray!75}624 & 601 & \cellcolor{gray!75}1247 & 1203 \\
F8  & \cellcolor{gray!75}701 & 524 & \cellcolor{gray!75}740.5 & 484.5 & \cellcolor{gray!75}1441.5 & 1008.5 \\
F9  & 612.5 & 612.5 & \cellcolor{gray!75}884.5 & 340.5 & \cellcolor{gray!75}1497 & 953 \\
F10 & \cellcolor{gray!75}676 & 549 & \cellcolor{gray!75}682.5 & 542.5 & \cellcolor{gray!75}1358.5 & 1091.5 \\
F11 & 572.5 & \cellcolor{gray!75}652.5 & 571.5 & \cellcolor{gray!75}653.5 & 1144 & \cellcolor{gray!75}1306 \\
F12 & \cellcolor{gray!75}624 & 601 & \cellcolor{gray!75}622 & 603 & \cellcolor{gray!75}1246 & 1204 \\
F13 & 575.5 & \cellcolor{gray!75}649.5 & 569 & \cellcolor{gray!75}656 & 1144.5 & \cellcolor{gray!75}1305.5 \\
F14 & \cellcolor{gray!75}699.5 & 525.5 & \cellcolor{gray!75}676.5 & 548.5 & \cellcolor{gray!75}1376 & 1074 \\
F15 & \cellcolor{gray!75}648 & 577 & \cellcolor{gray!75}647 & 578 & \cellcolor{gray!75}1295 & 1155 \\
F16 & \cellcolor{gray!75}646 & 579 & \cellcolor{gray!75}645 & 580 & \cellcolor{gray!75}1291 & 1159 \\
F17 & \cellcolor{gray!75}702 & 523 & \cellcolor{gray!75}701 & 524 & \cellcolor{gray!75}1403 & 1047 \\
F18 & 575 & \cellcolor{gray!75}650 & 554.5 & \cellcolor{gray!75}670.5 & 1129.5 & \cellcolor{gray!75}1320.5 \\
F19 & 551 & \cellcolor{gray!75}674 & 551 & \cellcolor{gray!75}674 & 1102 & \cellcolor{gray!75}1348 \\
F20 & \cellcolor{gray!75}637.5 & 587.5 & \cellcolor{gray!75}613 & 612 & \cellcolor{gray!75}1250.5 & 1199.5 \\
F21 & 597 & \cellcolor{gray!75}628 & 590.5 & \cellcolor{gray!75}634 & 1187.5 & \cellcolor{gray!75}1262.5 \\
F22 & 612.5 & 612.5 & \cellcolor{gray!75}905 & 320 & \cellcolor{gray!75}1517.5 & 932.5 \\
F23 & \cellcolor{gray!75}680 & 545 & \cellcolor{gray!75}681 & 544 & \cellcolor{gray!75}1361 & 1089 \\
F24 & 602.5 & \cellcolor{gray!75}622.5 & 603 & \cellcolor{gray!75}622 & 1205.5 & \cellcolor{gray!75}1244.5 \\
F25 & \cellcolor{gray!75}652 & 573 & \cellcolor{gray!75}641.5 & 583.5 & \cellcolor{gray!75}1293.5 & 1156.5 \\
F26 & 560 & \cellcolor{gray!75}665 & 567.5 & \cellcolor{gray!75}657.5 & 1127.5 & \cellcolor{gray!75}1322.5 \\
F27 & 594 & \cellcolor{gray!75}631 & 608 & \cellcolor{gray!75}617 & 1202 & \cellcolor{gray!75}1248 \\
F28 & \cellcolor{gray!75}625 & 600 & \cellcolor{gray!75}765.5 & 459.5 & \cellcolor{gray!75}1390.5 & 1059.5 \\
F29 & \cellcolor{gray!75}698 & 527 & 595 & \cellcolor{gray!75}630 & \cellcolor{gray!75}1293 & 1157 \\
F30 & \cellcolor{gray!75}700 & 525 & \cellcolor{gray!75}618.5 & 606.5 & \cellcolor{gray!75}1318.5 & 1131.5 \\
\hline
\textbf{Total} & \cellcolor{gray!75}\textbf{18275.5} & 17249.5 & \cellcolor{gray!75}\textbf{19172.5} & 16352.5 & \cellcolor{gray!75}\textbf{37448} & 33602 \\
\hline
\end{tabular}%
}
\end{table}

\subsection{Comparison with Competitive Algorithms}
\label{subsec:overall_comparison_ucop}
To further evaluate the competitiveness of DE-2LS, we compare it with several well-known and competitive algorithms, namely RDEx, LSRTDE, BlockEA, mLSHADE, jSO, and IEACOP. Table~\ref{tab:overall_algorithm_summary_ucop} reports the accuracy score, speed score, U-score, U-score rank sum, and average rank by mean final error. Since the U-score jointly evaluates both final accuracy and convergence speed, it provides a more comprehensive performance indicator than final error alone.

\begin{table}[!h]
\centering
\caption{Overall comparison of DE-2LS with competitive algorithms on UCOPs.}
\label{tab:overall_algorithm_summary_ucop}
\resizebox{\columnwidth}{!}{
\begin{tabular}{lrrrccc}
\toprule
Algos & Accuracy & Speed & U-score & Rank Sum & Avg. Rank & Relative Gap to DE-2LS \\
\midrule
\cellcolor{gray!25}DE-2LS  &\cellcolor{gray!25} \textbf{89113} &\cellcolor{gray!25} \textbf{89853.5} &\cellcolor{gray!25} \textbf{178966.5} \cellcolor{gray!25}& \textbf{61} &\cellcolor{gray!25} \textbf{2.43} & baseline \\
RDEx    & 87046.5 & 86420 & 173466.5 & 75 & 2.59 & $3.17\%$ \\
LSRTDE  & 84984 & 84049.5 & 169033.5 & 79 & 2.62 & $5.55\%$ \\
BlockEA & 52356 & 49498 & 101854 & 135 & 4.79 & $43.09\%$ \\
mLSHADE & 53952.5 & 45709.5 & 99662  & 137 & 4.55 & $44.31\%$ \\
jSO     & 54338.5 & 41939 & 96277.5  & 148 & 4.53 & $46.20\%$ \\
IEACOP  & 19734.5 & 44055.5 & 63790  & 177 & 6.48 & $64.36\%$ \\
\bottomrule
\end{tabular}}
\end{table}

As shown in Table~\ref{tab:overall_algorithm_summary_ucop}, DE-2LS achieves the best overall performance among all compared algorithms. It obtains the highest U-score of $178966.5$, along with the highest accuracy and speed scores. Compared with the RDEx baseline, DE-2LS improves the U-score by $3.17\%$. It also outperforms LSRTDE, which is the third-ranked method, by $5.55\%$. These results indicate that the proposed late-stage local-search mechanism improves not only the final objective error but also the overall convergence behavior of RDEx.

\begin{table*}[!h]
\centering
\caption{Function-wise pairwise U-score comparison of DE-2LS with competitive algorithms on UCOPs.}
\label{tab:functionwise_competitive_ucop}
\tiny
\setlength{\tabcolsep}{1.6pt}
\renewcommand{\arraystretch}{1.05}
\resizebox{\textwidth}{!}{%
\begin{tabular}{c|rrrrrrr|rrrrrrr|rrrrrrr}
\hline
\multirow{2}{*}{Func} &
\multicolumn{7}{c|}{Accuracy} &
\multicolumn{7}{c|}{Speed} &
\multicolumn{7}{c}{U-score} \\
\cline{2-22}
& LSRTDE & mLSHADE & jSO & BlockEA & IEACOP & RDEx & DE-2LS
& LSRTDE & mLSHADE & jSO & BlockEA & IEACOP & RDEx & DE-2LS
& LSRTDE & mLSHADE & jSO & BlockEA & IEACOP & RDEx & DE-2LS \\
\hline
F1 & 2537.5 & 2537.5 & 2537.5 & 2231 & 306.5 & 2537.5 & 2537.5 & 2306.5 & 1231 & 1664.5 & 300 & 3052 & 3036.5 & \cellcolor{gray!75}3634.5 & 4844 & 3768.5 & 4202 & 2531 & 3358.5 & 5574 & \cellcolor{gray!75}6172 \\
F3 & 2750 & 2750 & 2750 & 767.5 & 707.5 & 2750 & 2750 & 2575 & 1223.5 & 1934 & 300 & 2130.5 & 3264 & \cellcolor{gray!75}3798 & 5325 & 3973.5 & 4684 & 1067.5 & 2838 & 6014 & \cellcolor{gray!75}6548 \\
F4 & 325 & 3192 & 950 & \cellcolor{gray!75}3550 & 3383 & 2088 & 1737 & 2852 & 2142 & 1803.5 & 425 & \cellcolor{gray!75}3992 & 2157 & 1853.5 & 3177 & 5334 & 2753.5 & 3975 & \cellcolor{gray!75}7375 & 4245 & 3590.5 \\
F5 & \cellcolor{gray!75}3581.5 & 2112 & 1603 & 970 & 369 & 3220 & 3369.5 & 2803.5 & 1317.5 & 810 & \cellcolor{gray!75}3508.5 & 828.5 & 2799.5 & 3157.5 & 6385 & 3429.5 & 2413 & 4478.5 & 1197.5 & 6019.5 & \cellcolor{gray!75}6527 \\
F6 & 2722.5 & 2817.5 & 3037.5 & 1135 & 328 & 2147 & 3037.5 & 1667.5 & 3470.5 & 2830.5 & \cellcolor{gray!75}3691.5 & 313 & 1660.5 & 1591.5 & 4390 & \cellcolor{gray!75}6288 & 5868 & 4826.5 & 641 & 3807.5 & 4629 \\
F7 & 2489 & 1565 & 1291 & \cellcolor{gray!75}4050 & 351 & 2749 & 2730 & 1845 & 1170.5 & 817 & \cellcolor{gray!75}4050 & 1407.5 & 2956 & 2979 & 4334 & 2735.5 & 2108 & \cellcolor{gray!75}8100 & 1758.5 & 5705 & 5709 \\
F8 & 3428.5 & 2102 & 1561 & 888 & 437 & 3272.5 & \cellcolor{gray!75}3536 & 2756.5 & 1315.5 & 753.5 & 2743 & 1326.5 & 2969 & \cellcolor{gray!75}3361 & 6185 & 3417.5 & 2314.5 & 3631 & 1763.5 & 6241.5 & \cellcolor{gray!75}6897 \\
F9 & 2675 & 2675 & 2675 & 528.5 & 1321.5 & 2675 & 2675 & \cellcolor{gray!75}4038 & 3369 & 1249.5 & 1164 & 437 & 2208.5 & 2759 & \cellcolor{gray!75}6713 & 6044 & 3924.5 & 1692.5 & 1758.5 & 4883.5 & 5434 \\
F10 & 2853.5 & 1171 & 1266 & \cellcolor{gray!75}4044 & 340 & 2678.5 & 2872 & 2324.5 & 926 & 562 & \cellcolor{gray!75}4047 & 2488 & 2338 & 2539.5 & 5178 & 2097 & 1828 & \cellcolor{gray!75}8091 & 2828 & 5016.5 & 5411.5 \\
F11 & \cellcolor{gray!75}3688.5 & 1812 & 2252 & 300 & 999 & 3174.5 & 2999 & \cellcolor{gray!75}3263.5 & 1526.5 & 1518.5 & 300 & 2346.5 & 3215.5 & 3054.5 & \cellcolor{gray!75}6952 & 3338.5 & 3770.5 & 600 & 3345.5 & 6390 & 6053.5 \\
F12 & 3297.5 & 720 & 1658 & 2127 & 505 & 3442 & \cellcolor{gray!75}3475.5 & 3181 & 353 & 1266 & 2126 & 1477 & 3390 & \cellcolor{gray!75}3432 & 6478.5 & 1073 & 2924 & 4253 & 1982 & 6832 & \cellcolor{gray!75}6907.5 \\
F13 & 3214.5 & 2009 & 1868 & 629 & 692 & \cellcolor{gray!75}3469.5 & 3343 & 3190 & 1898 & 1760 & 440 & 1196 & \cellcolor{gray!75}3448 & 3293 & 6404.5 & 3907 & 3628 & 1069 & 1888 & \cellcolor{gray!75}6917.5 & 6636 \\
F14 & 3003 & 1648 & 2540 & 925 & 300 & 3277.5 & \cellcolor{gray!75}3531.5 & 3113.5 & 1615.5 & 2214.5 & 300 & 980 & 3409.5 & \cellcolor{gray!75}3592 & 6116.5 & 3263.5 & 4754.5 & 1225 & 1280 & 6687 & \cellcolor{gray!75}7123.5 \\
F15 & 2942 & 1553 & 2195 & 684 & 627 & 3557 & \cellcolor{gray!75}3667 & 2902 & 1534.5 & 2126 & 323.5 & 1129 & 3557 & \cellcolor{gray!75}3653 & 5844 & 3087.5 & 4321 & 1007.5 & 1756 & 7114 & \cellcolor{gray!75}7320 \\
F16 & \cellcolor{gray!75}3404 & 1934 & 1455 & 1428 & 346 & 3273 & 3385 & \cellcolor{gray!75}3370.5 & 1860.5 & 1393.5 & 799 & 1211.5 & 3243.5 & 3346.5 & \cellcolor{gray!75}6774.5 & 3794.5 & 2848.5 & 2227 & 1557.5 & 6516.5 & 6731.5 \\
F17 & 2969 & 1657 & 1839 & 1322 & 470 & 3335 & \cellcolor{gray!75}3633 & 2990.5 & 1565.5 & 1322.5 & 939 & 1419 & 3355 & \cellcolor{gray!75}3633.5 & 5959.5 & 3222.5 & 3161.5 & 2261 & 1889 & 6690 & \cellcolor{gray!75}7266.5 \\
F18 & 2484 & 1291 & 2767 & 1472 & 327 & \cellcolor{gray!75}3485 & 3399 & 2947 & 1398.5 & 2433 & 1080.5 & 350 & \cellcolor{gray!75}3569.5 & 3446.5 & 5431 & 2689.5 & 5200 & 2552.5 & 677 & \cellcolor{gray!75}7054.5 & 6845.5 \\
F19 & \cellcolor{gray!75}3642 & 1310 & 2133 & 1136 & 382 & 3411 & 3211 & \cellcolor{gray!75}3592 & 1367 & 2032.5 & 492 & 1158.5 & 3390 & 3193 & \cellcolor{gray!75}7234 & 2677 & 4165.5 & 1628 & 1540.5 & 6801 & 6404 \\
F20 & \cellcolor{gray!75}3462.5 & 1226 & 1679 & 2293 & 357 & 3057.5 & 3150 & \cellcolor{gray!75}3433 & 1494 & 1596.5 & 770.5 & 1399.5 & 3232.5 & 3299 & \cellcolor{gray!75}6895.5 & 2720 & 3275.5 & 3063.5 & 1756.5 & 6290 & 6449 \\
F21 & 2972 & 1577 & 1013 & \cellcolor{gray!75}3721 & 327 & 2832 & 2783 & 2764.5 & 1186.5 & 626.5 & \cellcolor{gray!75}3722.5 & 1306 & 2845 & 2774 & 5736.5 & 2763.5 & 1639.5 & \cellcolor{gray!75}7443.5 & 1633 & 5677 & 5557 \\
F22 & 2800 & 2800 & 2800 & 300 & 925 & 2800 & 2800 & 1625 & 3200 & 925 & 300 & \cellcolor{gray!75}4050 & 2270 & 2855 & 4425 & \cellcolor{gray!75}6000 & 3725 & 600 & 4975 & 5070 & 5655 \\
F23 & 2960.5 & 1542 & 1732 & 1800 & 707 & 3128.5 & \cellcolor{gray!75}3355 & 2970.5 & 1774 & 1198.5 & 1800 & 928.5 & 3170 & \cellcolor{gray!75}3383.5 & 5931 & 3316 & 2930.5 & 3600 & 1635.5 & 6298.5 & \cellcolor{gray!75}6738.5 \\
F24 & 2902 & 1498 & 1089 & \cellcolor{gray!75}4050 & 300 & 2698.5 & 2687.5 & 2828.5 & 1546.5 & 546 & \cellcolor{gray!75}4022 & 869.5 & 2707.5 & 2705 & 5730.5 & 3044.5 & 1635 & \cellcolor{gray!75}8072 & 1169.5 & 5406 & 5392.5 \\
F25 & 1764 & 2781 & 1661 & 300 & 1443 & 3588 & \cellcolor{gray!75}3688 & 2535.5 & 1369.5 & 2138.5 & 300 & 1495.5 & 3672.5 & \cellcolor{gray!75}3713.5 & 4299.5 & 4150.5 & 3799.5 & 600 & 2938.5 & 7260.5 & \cellcolor{gray!75}7401.5 \\
F26 & 2523.5 & 916 & 1213 & \cellcolor{gray!75}3850 & 1240 & 2816 & 2666.5 & 2485 & 1400 & 725 & \cellcolor{gray!75}3850 & 1250 & 2827.5 & 2687.5 & 5008.5 & 2316 & 1938 & \cellcolor{gray!75}7700 & 2490 & 5643.5 & 5354 \\
F27 & 2830 & 1251 & 1593 & 1461 & 645 & \cellcolor{gray!75}3732 & 3713 & 3181 & 975.5 & 1731.5 & 1634.5 & 842 & \cellcolor{gray!75}3454.5 & 3406 & 6011 & 2226.5 & 3324.5 & 3095.5 & 1487 & \cellcolor{gray!75}7186.5 & 7119 \\
F28 & 2837.5 & 2837.5 & 2663.5 & 300 & 997 & 2752 & 2837.5 & 2507 & 994 & 1625 & 300 & \cellcolor{gray!75}3665.5 & 2863.5 & 3270 & 5344.5 & 3831.5 & 4288.5 & 600 & 4662.5 & 5615.5 & \cellcolor{gray!75}6107.5 \\
F29 & \cellcolor{gray!75}4050 & 1399 & 1309 & 2887 & 302 & 2542 & 2736 & \cellcolor{gray!75}4048 & 1271.5 & 1152.5 & 2561.5 & 552 & 2839.5 & 2800 & \cellcolor{gray!75}8098 & 2670.5 & 2461.5 & 5448.5 & 854 & 5381.5 & 5536 \\
F30 & \cellcolor{gray!75}3875 & 1269 & 1208 & 3207 & 300 & 2558 & 2808 & \cellcolor{gray!75}3953 & 1213.5 & 1183 & 3208 & 454.5 & 2570.5 & 2642.5 & \cellcolor{gray!75}7828 & 2482.5 & 2391 & 6415 & 754.5 & 5128.5 & 5450.5 \\
\hline
\textbf{Total} & 84984 & 53952.5 & 54338.5 & 52356 & 19734.5 & 87046.5 & \cellcolor{gray!75}\textbf{89113} & 84049.5 & 45709.5 & 41939 & 49498 & 44055.5 & 86420 & \cellcolor{gray!75}\textbf{89853.5} & 169033.5 & 99662 & 96277.5 & 101854 & 63790 & 173466.5 & \cellcolor{gray!75}\textbf{178966.5} \\
\hline
\end{tabular}%
}
\end{table*}

\begin{table*}[!h]
\centering
\caption{Mean and standard deviation comparison of other competitive algorithms.}\label{tab:mean_std_comparison}
\scriptsize
\setlength{\tabcolsep}{2.5pt}\resizebox{1\textwidth}{!}{\begin{tabular}{c|c|c|c|c|c|c|c}\hline
\multirow{2}{*}{Func}	&	\multicolumn{1}{c}{	LSRTDE	}		&	\multicolumn{1}{c}{	mLSHADE\_LR	}		&	\multicolumn{1}{c}{	jSO 	}		&	\multicolumn{1}{c}{	BlockEA	}		&	\multicolumn{1}{c}{IEACOP} 			&	\multicolumn{1}{c}{	RDEx 	}		&	\multicolumn{1}{c}{	DE-2LS 	}		\\
\cline{2-8}	&		Mean	$\pm$	Std	&		Mean	$\pm$	Std	&		Mean	$\pm$	Std	&		Mean	$\pm$	Std	&	Mean	$\pm$	Std	&		Mean	$\pm$	Std	&		Mean	$\pm$	Std	\\\hline
F1	&	\cellcolor{gray!75}	0.00E+00	$\pm$	0.00E+00	&	\cellcolor{gray!75}	0.00E+00	$\pm$	0.00E+00	&		4.55E-15	$\pm$	6.77E-15	&		1.13E-07	$\pm$	5.50E-07	&	2.85E-06	$\pm$	8.34E-07	&	\cellcolor{gray!75}	0.00E+00	$\pm$	0.00E+00	&	\cellcolor{gray!75}	0.00E+00	$\pm$	0.00E+00	\\
F3	&	\cellcolor{gray!75}	0.00E+00	$\pm$	0.00E+00	&		5.00E-14	$\pm$	2.50E-14	&		4.32E-14	$\pm$	2.48E-14	&		2.60E-04	$\pm$	1.23E-03	&	1.20E-06	$\pm$	1.72E-06	&	\cellcolor{gray!75}	0.00E+00	$\pm$	0.00E+00	&	\cellcolor{gray!75}	0.00E+00	$\pm$	0.00E+00	\\
F4	&		5.88E+01	$\pm$	1.11E+00	&		6.93E+00	$\pm$	1.52E+01	&		5.86E+01	$\pm$	2.18E-14	&	\cellcolor{gray!75}	9.27E-03	$\pm$	1.08E-02	&	1.91E+00	$\pm$	2.03E+00	&		1.93E+01	$\pm$	2.04E+00	&		2.00E+01	$\pm$	2.21E+00	\\
F5	&	\cellcolor{gray!75}	2.31E+00	$\pm$	9.83E-01	&		8.08E+00	$\pm$	3.14E+00	&		1.07E+01	$\pm$	2.21E+00	&		1.87E+01	$\pm$	9.71E+00	&	2.99E+01	$\pm$	7.46E+00	&		3.42E+00	$\pm$	1.93E+00	&		2.87E+00	$\pm$	1.56E+00	\\
F6	&		3.50E-08	$\pm$	7.94E-08	&		2.95E-03	$\pm$	1.05E-02	&		1.23E-13	$\pm$	3.15E-14	&		7.04E-05	$\pm$	1.67E-04	&	1.32E-01	$\pm$	2.09E-01	&		1.60E-07	$\pm$	2.75E-07	&	\cellcolor{gray!75}	0.00E+00	$\pm$	0.00E+00	\\
F7	&		3.81E+01	$\pm$	7.69E+00	&		3.98E+01	$\pm$	3.17E+00	&		4.07E+01	$\pm$	2.16E+00	&	\cellcolor{gray!75}	6.34E-06	$\pm$	1.23E-05	&	5.39E+01	$\pm$	5.87E+00	&		3.57E+01	$\pm$	1.46E+00	&		3.59E+01	$\pm$	2.36E+00	\\
F8	&		2.35E+00	$\pm$	1.52E+00	&		7.95E+00	$\pm$	2.66E+00	&		1.11E+01	$\pm$	2.05E+00	&		2.01E+01	$\pm$	8.02E+00	&	2.75E+01	$\pm$	6.46E+00	&		2.71E+00	$\pm$	1.17E+00	&	\cellcolor{gray!75}	2.07E+00	$\pm$	1.28E+00	\\
F9	&	\cellcolor{gray!75}	0.00E+00	$\pm$	0.00E+00	&	\cellcolor{gray!75}	0.00E+00	$\pm$	0.00E+00	&	\cellcolor{gray!75}	0.00E+00	$\pm$	0.00E+00	&		1.24E+00	$\pm$	1.15E+00	&	2.38E+00	$\pm$	9.93E+00	&	\cellcolor{gray!75}	0.00E+00	$\pm$	0.00E+00	&	\cellcolor{gray!75}	0.00E+00	$\pm$	0.00E+00	\\
F10	&		2.88E+02	$\pm$	1.62E+02	&		1.47E+03	$\pm$	2.92E+02	&		1.40E+03	$\pm$	2.42E+02	&	\cellcolor{gray!75}	1.38E+00	$\pm$	2.92E+00	&	2.64E+03	$\pm$	4.76E+02	&		3.55E+02	$\pm$	1.99E+02	&		3.13E+02	$\pm$	2.41E+02	\\
F11	&	\cellcolor{gray!75}	1.99E-01	$\pm$	4.97E-01	&		8.12E+00	$\pm$	1.10E+01	&		7.38E+00	$\pm$	1.66E+01	&		1.11E+03	$\pm$	4.93E+01	&	3.16E+01	$\pm$	1.36E+01	&		1.39E+00	$\pm$	1.80E+00	&		1.75E+00	$\pm$	1.94E+00	\\
F12	&		1.28E+00	$\pm$	1.61E+00	&		1.18E+03	$\pm$	4.34E+02	&		1.74E+02	$\pm$	9.70E+01	&		2.95E+01	$\pm$	2.87E+01	&	1.50E+03	$\pm$	4.86E+02	&	\cellcolor{gray!75}	1.09E+00	$\pm$	1.50E+00	&		1.10E+00	$\pm$	1.83E+00	\\
F13	&		9.79E+00	$\pm$	6.72E+00	&		1.97E+01	$\pm$	8.52E+00	&		1.95E+01	$\pm$	2.50E+00	&		7.13E+02	$\pm$	9.93E+02	&	1.58E+02	$\pm$	1.95E+02	&	\cellcolor{gray!75}	6.49E+00	$\pm$	6.78E+00	&		9.04E+00	$\pm$	6.69E+00	\\
F14	&		1.06E+01	$\pm$	1.04E+01	&		2.28E+01	$\pm$	3.42E+00	&		1.24E+01	$\pm$	9.45E+00	&		3.76E+01	$\pm$	5.69E+00	&	1.02E+02	$\pm$	2.80E+01	&		7.49E+00	$\pm$	9.86E+00	&	\cellcolor{gray!75}	6.49E+00	$\pm$	9.65E+00	\\
F15	&		1.11E+00	$\pm$	1.03E+00	&		1.25E+01	$\pm$	1.32E+01	&		3.27E+00	$\pm$	1.67E+00	&		1.89E+02	$\pm$	5.31E+02	&	7.86E+01	$\pm$	8.48E+01	&		4.68E-01	$\pm$	2.85E-01	&	\cellcolor{gray!75}	3.72E-01	$\pm$	9.89E-02	\\
F16	&	\cellcolor{gray!75}	2.99E+00	$\pm$	8.78E-01	&		5.57E+01	$\pm$	5.83E+01	&		7.83E+01	$\pm$	8.70E+01	&		6.96E+01	$\pm$	8.05E+01	&	3.75E+02	$\pm$	1.52E+02	&		4.70E+00	$\pm$	3.38E+00	&		3.96E+00	$\pm$	2.69E+00	\\
F17	&		2.45E+01	$\pm$	2.42E+00	&		3.60E+01	$\pm$	9.18E+00	&		3.36E+01	$\pm$	8.29E+00	&		6.35E+01	$\pm$	5.05E+01	&	1.08E+02	$\pm$	6.84E+01	&	\cellcolor{gray!75}	2.03E+01	$\pm$	8.18E+00	&		2.04E+01	$\pm$	6.01E+00	\\
F18	&		1.76E+01	$\pm$	7.61E+00	&		3.05E+01	$\pm$	7.56E+00	&		1.84E+01	$\pm$	6.40E+00	&		4.98E+01	$\pm$	5.99E+01	&	2.28E+02	$\pm$	5.81E+01	&	\cellcolor{gray!75}	1.26E+01	$\pm$	1.01E+01	&		1.50E+01	$\pm$	9.01E+00	\\
F19	&	\cellcolor{gray!75}	1.63E+00	$\pm$	1.76E-01	&		1.09E+01	$\pm$	4.88E+00	&		4.81E+00	$\pm$	1.38E+00	&		3.38E+01	$\pm$	8.34E+01	&	7.72E+01	$\pm$	7.39E+01	&		1.97E+00	$\pm$	5.53E-01	&		2.16E+00	$\pm$	5.52E-01	\\
F20	&	\cellcolor{gray!75}	2.43E+00	$\pm$	5.93E+00	&		4.15E+01	$\pm$	1.04E+01	&		2.99E+01	$\pm$	6.89E+00	&		6.70E+01	$\pm$	1.88E+02	&	1.84E+02	$\pm$	3.18E+01	&		1.60E+01	$\pm$	3.81E+01	&		1.16E+01	$\pm$	2.43E+01	\\
F21	&		2.02E+02	$\pm$	1.92E+00	&		2.08E+02	$\pm$	2.02E+00	&		2.11E+02	$\pm$	2.10E+00	&	\cellcolor{gray!75}	5.94E+01	$\pm$	7.35E+01	&	2.24E+02	$\pm$	6.99E+00	&		2.03E+02	$\pm$	1.57E+00	&		2.03E+02	$\pm$	1.85E+00	\\
F22	&	\cellcolor{gray!75}	1.00E+02	$\pm$	0.00E+00	&	\cellcolor{gray!75}	1.00E+02	$\pm$	0.00E+00	&	\cellcolor{gray!75}	1.00E+02	$\pm$	3.30E-13	&		2.18E+02	$\pm$	6.60E+00	&	1.00E+02	$\pm$	1.59E-08	&	\cellcolor{gray!75}	1.00E+02	$\pm$	0.00E+00	&	\cellcolor{gray!75}	1.00E+02	$\pm$	0.00E+00	\\
F23	&		3.41E+02	$\pm$	5.50E+00	&		3.57E+02	$\pm$	7.50E+00	&		3.54E+02	$\pm$	4.05E+00	&		1.17E+03	$\pm$	9.36E+02	&	3.77E+02	$\pm$	1.19E+01	&		3.40E+02	$\pm$	4.13E+00	&	\cellcolor{gray!75}	3.38E+02	$\pm$	3.17E+00	\\
F24	&		4.14E+02	$\pm$	5.16E+00	&		4.25E+02	$\pm$	4.44E+00	&		4.28E+02	$\pm$	2.65E+00	&	\cellcolor{gray!75}	3.67E+02	$\pm$	9.66E+00	&	4.42E+02	$\pm$	3.97E+00	&		4.16E+02	$\pm$	5.30E+00	&		4.16E+02	$\pm$	3.30E+00	\\
F25	&		3.87E+02	$\pm$	2.98E-03	&		3.81E+02	$\pm$	2.67E+00	&		3.87E+02	$\pm$	6.81E-03	&		4.44E+02	$\pm$	1.03E+01	&	3.86E+02	$\pm$	1.68E+00	&		3.79E+02	$\pm$	8.06E-01	&	\cellcolor{gray!75}	3.79E+02	$\pm$	7.65E-01	\\
F26	&		6.63E+02	$\pm$	1.45E+02	&		9.91E+02	$\pm$	7.57E+01	&		9.43E+02	$\pm$	3.61E+01	&		3.87E+02	$\pm$	1.69E+00	&	1.06E+03	$\pm$	4.82E+02	&		5.87E+02	$\pm$	1.60E+02	&	\cellcolor{gray!75}	6.34E+02	$\pm$	1.47E+02	\\
F27	&		4.71E+02	$\pm$	3.31E+00	&		5.04E+02	$\pm$	9.06E+00	&		5.00E+02	$\pm$	5.10E+00	&		5.02E+02	$\pm$	1.37E+01	&	5.12E+02	$\pm$	6.90E+00	&	\cellcolor{gray!75}	4.61E+02	$\pm$	4.79E+00	&		4.61E+02	$\pm$	4.87E+00	\\
F28	&	\cellcolor{gray!75}	3.00E+02	$\pm$	0.00E+00	&	\cellcolor{gray!75}	3.00E+02	$\pm$	0.00E+00	&		3.09E+02	$\pm$	3.16E+01	&		5.09E+02	$\pm$	5.04E+00	&	3.19E+02	$\pm$	4.43E+01	&		3.04E+02	$\pm$	2.07E+01	&	\cellcolor{gray!75}	3.00E+02	$\pm$	0.00E+00	\\
F29	&	\cellcolor{gray!75}	2.80E+02	$\pm$	7.30E+00	&		4.27E+02	$\pm$	2.06E+01	&		4.31E+02	$\pm$	8.86E+00	&		3.42E+02	$\pm$	5.89E+01	&	5.19E+02	$\pm$	4.96E+01	&		3.90E+02	$\pm$	1.39E+01	&		3.81E+02	$\pm$	1.96E+01	\\
F30	&	\cellcolor{gray!75}	4.04E+02	$\pm$	2.25E+00	&		1.95E+03	$\pm$	1.69E+02	&		1.97E+03	$\pm$	1.54E+01	&		4.48E+02	$\pm$	6.93E+01	&	3.22E+03	$\pm$	5.55E+02	&		8.40E+02	$\pm$	4.31E+02	&		5.73E+02	$\pm$	2.30E+02	\\\hline
\end{tabular}}
\end{table*}

The rank-based indicators further support the superiority of DE-2LS. It achieves the smallest U-score rank sum of $61$, followed by RDEx and LSRTDE with rank sums of $75$ and $79$, respectively. Since a smaller rank sum indicates better and more consistent performance across functions, this result confirms the robustness of DE-2LS. DE-2LS also obtains the best average rank by mean final error, with a value of $2.43$, which shows that the proposed method remains highly competitive even when only final solution quality is considered.

Table~\ref{tab:functionwise_competitive_ucop} provides a more detailed function-wise comparison in terms of accuracy, speed, and U-score. The shaded cells indicate the unique best value in each metric group, while equal best values are intentionally left unshaded. DE-2LS achieves the best U-score on $11$ out of the $29$ functions, the largest number among all compared algorithms. LSRTDE achieves the best U-score on $7$ functions, BlockEA on $5$ functions, RDEx on $3$ functions, mLSHADE on $2$ functions, and IEACOP on $1$ function. This confirms that although several algorithms are competitive on specific functions, DE-2LS provides the strongest overall U-score behavior across the benchmark set.

A closer inspection of the two U-score components shows that DE-2LS has a clear speed advantage. It achieves the best speed score across $9$ functions, surpassing any other algorithm. This explains its highest overall speed score in Table~\ref{tab:overall_algorithm_summary_ucop}. The speed improvement is particularly visible on functions such as F1, F3, F8, F12, F14, F15, F17, and F25, where DE-2LS reaches competitive objective-error levels earlier than the other algorithms. This supports the intended role of the LS component as a budget-aware refinement mechanism that accelerates the best-so-far trajectory.

In terms of accuracy, DE-2LS also performs strongly. It achieves the best or tied-best accuracy on $13$ functions, indicating that the proposed LS component does not degrade the final solution quality of RDEx. Instead, it improves final-error performance across several functions while maintaining strong convergence speed. Table~\ref{tab:mean_std_comparison} further shows that DE-2LS achieves competitive mean errors across many functions and often matches or improves upon RDEx. Overall, these results demonstrate that DE-2LS provides a stronger balance between final accuracy and convergence speed than the other compared algorithms.

\section{Conclusion}
\label{sec:conclusion}

This paper presents DE-2LS, a late-stage, locally search-enhanced RDEx variant for unconstrained single-objective numerical optimization with variable bounds. The proposed method preserves the original RDEx evolutionary framework and strengthens it through two late-stage components: smoothed exploitation-biased branch-rate control and a guarded coordinate-pattern local-search module. A staged ablation study was conducted to determine the final configuration. In the pre-LS stage, the best setting was obtained by combining late EB-rate smoothing, refined standard-branch Gaussian sampling, and jointly tuned selection-pressure parameters. After fixing this configuration, the LS module was further analyzed through timing, budget, and step-size studies, where the timing-oriented setting with earlier LS activation achieved the best overall performance. The final DE-2LS variant consistently improves upon RDEx. In a direct head-to-head comparison, it increases the U-score from $33602.0$ to $37448.0$, corresponding to a gain of $11.45\%$. In the broader comparison with competitive algorithms, DE-2LS achieves the highest total U-score of $178966.5$, outperforming the other algorithms by $34.43\%$ in average. These results show that the proposed LS module works effectively as a budget-aware late-stage refinement mechanism, improving both convergence speed and final objective quality without replacing the main RDEx search engine. Future work will investigate more adaptive LS triggering rules and evaluate the proposed framework on broader benchmark suites and real-world continuous optimization problems.

\section*{Acknowledgment}
The authors sincerely thank Prof. P. N. Suganthan and the organizers of the IEEE CEC numerical optimization competition for providing the official U-score evaluation procedure and related competition resources. These materials were highly valuable for implementing the evaluation framework and ensuring consistency with the intended competition-based assessment protocol used in this work. The authors also thank the authors of RDEx for providing the source code, which facilitated the implementation of the baseline method and the development of the proposed DE-2LS algorithm.

\bibliographystyle{IEEEtran}
\bibliography{rdex_ls_references}

\end{document}